\documentclass[english,runningheads]{llncs}
\usepackage[T1]{fontenc}
\usepackage[latin9]{inputenc}
\usepackage{amsbsy}
\usepackage{graphicx}

\makeatletter

\providecommand{\tabularnewline}{\\}

\usepackage{cite}
\usepackage{amsmath,amssymb,amsfonts}
\usepackage{algorithmic}
\usepackage{graphicx}
\usepackage{textcomp}
\usepackage{xcolor}
\usepackage{float}
\usepackage{array}
\newcolumntype{H}{>{\setbox0=\hbox\bgroup}c<{\egroup}@{}}

\@ifundefined{showcaptionsetup}{}{%
 \PassOptionsToPackage{caption=false}{subfig}}
\usepackage{subfig}
\makeatother

\usepackage{babel}
\begin{document}

\title{CascadeML: An Automatic Neural Network Architecture Evolution and Training Algorithm for Multi-label Classification\thanks{This research was supported by Science Foundation Ireland (SFI) under Grant Number SFI/12/RC/2289.}}
\titlerunning{CascadeML}

%
%
\author{Arjun Pakrashi\and
Brian Mac\ Namee
}
\authorrunning{Pakrashi and Mac\ Namee}
%
\institute{Insight Centre for Data Analytics \\
University College Dublin\\
Dublin, Ireland
\email{arjun.pakrashi@ucdconnect.ie,arjun.pakrashi@insight-centre.org}\\
\email{brian.macnamee@ucd.ie}}

\maketitle
\begin{abstract}
Multi-label classification is an approach which allows a datapoint
to be labelled with more than one class at the same time. A common
but trivial approach is to train individual binary classifiers per
label, but the performance can be improved by considering associations
within the labels. Like with any machine learning algorithm, hyperparameter
tuning is important to train a good multi-label classifier model.
The task of selecting the best hyperparameter settings for an algorithm is
an optimisation problem. Very limited work has been done on automatic
hyperparameter tuning and AutoML in the multi-label domain. This paper
attempts to fill this gap by proposing a neural network algorithm,
CascadeML, to train multi-label neural network based on cascade neural
networks. This method requires minimal or no hyperparameter tuning
and also considers pairwise label associations. The cascade algorithm
grows the network architecture incrementally in a two phase process
as it learns the weights using adaptive first order gradient algorithm,
therefore omitting the requirement of preselecting the number of hidden
layers, nodes and the learning rate. The method was tested on 10 multi-label
datasets and compared with other multi-label classification algorithms.
Results show that CascadeML performs very well without hyperparameter
tuning.
\end{abstract}

\section{Introduction}

In \emph{multi-label classification} problems a datapoint can be assigned to more than one class, or \emph{label}, simultaneously \cite{DBLP:books/sp/HerreraCRJ16}. For example, an image can be classified as containing multiple different objects, or music can be labelled with more than one genre. This contrasts with  \emph{multi-class classification} problems in which objects can only belong to a single class. Multi-label classification algorithms either break the multi-label problem down into smaller multi-class classification problems---for example \emph{classifier chains} \cite{Read2011}---and are known as \emph{problem transformation} methods---or modify multi-class algorithms to directly train on multi-label datasets---for example \emph{BackPropagation in Multi-Label Learning} (BPMLL) \cite{Zhang:2006:MNN:1159162.1159294}---and are known as \emph{algorithm adaptation} methods. 

\emph{Automatic machine learning} \cite{NIPS2015_5872}, or \emph{AutoML},  approaches have seen a recent resurgence of interest as researchers look for ways to automatically select optimal algorithms, features, model architectures, and hyperparameters for machine learning tasks. The AutoML research community has, however, paid very little attention to multi-label classification problems, although there have been some recent efforts \cite{deSa:2017:TMA:3067695.3082053,kar68970,DBLP:journals/corr/abs-1811-04060}.

The \emph{Cascade2} algorithm \cite{PRECHELT1997885} is an interesting neural network approach that learns model parameters and model architecture at the same time. In Cascade2, which is based on the cascade correlation neural network approach \cite{NIPS1989_207}, training starts with a simple perceptron network, which is grown incrementally by adding new cascaded layers with skip-level connections as long as performance on a validation dataset improves. Weights in each new layer are trained independently of the overall network which greatly reduces the processing burden of this approach.

This paper proposes \emph{CascadeML}, a new AutoML solution for multi-label classification problems, that is inspired by the Cascade2 algorithm and BPMLL. Improvements are made to both components leading to an implementation that requires minimal hyperparameter or network architecture tuning. In a series of evaluation experiments this approach has been shown to perform very well without the extensive hyperparameter
tuning required by state-of-the-art multi-label classification methods. To the best of authors' knowledge this is the first automatic neural network architecture selection and training approach for multi-label classification methods.

The remainder of the paper is structured as follows. Section \ref{sec:Literature-Review} discusses the existing literature including a formal definition of multi-label classification and the BPMLL algorithm. Section \ref{sec:Cascade-Correlation-Neural} describes the cascade neural network approach and, specifically, the Cascade2 algorithm. The proposed CascadeML method is then presented in Section \ref{sec:Proposed-Method:-CascadeML}.
The design of an experiment to evaluate the performance of the CascadeML algorithm, and benchmarking its performance against state-of-the-art multi-label classification approaches is described in Section \ref{sec:Experiment}. Section \ref{sec:Results} presents and discusses the results of this experiment.
Finally, Section \ref{sec:Conclusion-and-Future} discusses future
research directions and concludes the paper.

\section{Related Work}\label{sec:Literature-Review}

In this section first the cost function of BPMLL will be mentioned followed by a brief review of AutoML in multi-label literature. Then the Cascade2 algorithm will be explained.

\subsection{BPMLL}
The first neural-network based multi-label algorithm, BackPropagation
in Multi-Label Learning (BPMLL), was proposed by Zhang et al. in
2006 \cite{Zhang:2006:MNN:1159162.1159294}. It is a single hidden
layer, fully connected feed-forward architecture, which uses the back-propagation of error algorithm to optimise a
variation of the ranking loss function \cite{zhang2014review} that takes pairwise label associations into account. This loss function can be defined as follows:

\begin{equation}
E=\sum_{i=1}^{n}\frac{1}{|\boldsymbol{y}_{i}||\bar{\boldsymbol{y}}_{i}|}\sum_{(k,l)\in(\boldsymbol{y}_{i}\times\bar{\boldsymbol{y}}_{i})}exp(-(c_{i}^{(k)}-c_{i}^{(l)}))\label{eq:bpmll_cost}
\end{equation}

\noindent Here $\boldsymbol{y}_{i}$ indicates the set of 
labels assigned to $\boldsymbol{x}_{i}$ and $\bar{\boldsymbol{y}}_{i}$
indicates the set of labels which are not assigned to $\boldsymbol{x}_{i}$. The network uses the $tanh$ activation function, therefore this algorithm
uses a bipolar encoding of the target variables: $y_{i}^{(l)}=+1$
if the label $l$ is relevant to $\boldsymbol{x}_{i}$, and
$-1$ if irrelevant. Here $c_{i}^{(k)}$ and $c_{i}^{(l)}$
are the outputs of the $k^{th}$ and the $l^{th}$ output units representing the corresponding label predictions for the datapoint $\boldsymbol{x}_{i}$. 

The intuition behind this loss function is that for a pair of labels $(k,l)$, where $k$ is relevant to the datapoint $\boldsymbol{x}_{i}$ and $l$ is not, if the prediction score for $k$ is positive whereas the prediction score for $l$ is negative, then $exp (-(c_{i}^{(k)}-c_{i}^{(l)}))$ has the minimum penalty. An incorrect prediction score order results in higher penalty. Therefore, minimising Eq. \eqref{eq:bpmll_cost} would result in pairs of labels being predicted correctly.

For BPMLL, like any neural network algorithm, the number of hidden units has to be determined, which is a hyperparameter to be tuned.
In \cite{10.1007/978-3-540-87700-4_41} modifications to the BPMLL
loss function were proposed. This modified version learns the network as in BPMLL,
and also learns the values using which the predicted scores are thresholded
to get label assignments. 

There have been a small number of other neural network approaches specifically designed for multi-label classification scenarios. In 2009, Zhang et al. proposed a \emph{multi-label-based radial basis
function} network, ML-RBF \cite{Zhang2009}. This is an extension
of the RBF network, optimising the sum-of-squares
function.
\emph{Multi-class multi-label perceptron}(MMP) \cite{Crammer:2003:FAO:944919.944962}
trains perceptrons for each label but in a way such that the applicable
labels are ranked higher than the incorrect labels, thus considering
associations between labels. An improvement of MMP, \emph{multi-label
pairwise perceptron} (MLPP), was proposed in \cite{4634206}. This  approach
trains the perceptrons for each pair of classes. Nam, et al. \cite{10.1007/978-3-662-44851-9_28}
demonstrate the efficiency and effectiveness of cross-entropy for multi-label classification,
improving the work of BPMLL by using several recent developments
such as ReLU activation units, dropout and the use of the adaptive gradient descent algorithm AdaGrad \cite{Goodfellow-et-al-2016}.

Some work involving deep neural networks on computer vision and image recognition were done in \cite{ZHUANG2018225,ZHU2017224,YU201738,CHEN2013298}, which uses multi-label datasets as a part of the training pipeline. Similarly, convolutional neural networks was extended to predict multi-label images in \cite{DBLP:journals/corr/WeiXHNDZY14}. In \cite{DBLP:journals/corr/ReadP15} the feature space of multi-label
classification was modified using deep belief networks such that the labels become less dependent, after which well-known multi-label algorithms are applied in the modified space. 

\subsection{AutoML}

AutoML algorithms focusing on multi-label specific problems are approached
in \cite{deSa:2017:TMA:3067695.3082053,kar68970}, using genetic algorithms to train and select multi-label models.
Wever et al. \cite{DBLP:journals/corr/abs-1811-04060} propose an
extension of an existing multi-class AutoML tool for multi-label.
Except these works, no other AutoML based or automatic hyperparameter
tuning based work on the multi-label domain was found.

The cascade correlation neural network approach \cite{NIPS1989_207} was an early AutoML method. In cascade correlation neural networks training starts with a simple perceptron network, which is grown incrementally by adding new cascaded layers with skip-level connections as long as performance on a validation dataset improves. Since the proposal of the original cascade correlation algorithm in
\cite{NIPS1989_207}, various improvements that follow a similar overall process to the original method have been proposed, for example in \cite{Baluja-1994-13783,10.1007/978-1-4471-2097-1_189,waugh1994connection,329690}, as well as Cascade2 \cite{PRECHELT1997885}. Active research in this field, however, is fairly limited. As it is the basis for CascadeML, the Cascade2 algorithm is described in detail in the next section.

\subsection{The Cascade2 Algorithm\label{sec:Cascade-Correlation-Neural}}

This section describes the Casecade2 algorithm upon which CascadeML is based. The generic architecture of a cascade neural network is first described, before the specific Casecade2 training algorithm is presented. 

\subsubsection{Architecture\label{subsec:Architecture}}

The \emph{cascade correlation neural network}, first proposed by Fahlman \& Lebiere \cite{NIPS1989_207},
is an incremental greedy multi-class neural network learning algorithm which
grows the network architecture at the same time as it trains the network weights. For a multi-class classification problem with $d$ inputs and $q$ classes, the architecture of a network trained using CascadeML will have $d+1$ inputs (including a bias term) and $q$ outputs (one for each
class). Each of the network's $L$ hidden layers, $l_{i}$, will have only one unit, which receives
incoming weights from all the $d+1$ inputs as well as from all the hidden units
in the previous layers. The output of each hidden layer $l_{i}$ is
connected to the $q$ outputs of the network. A layer with such a
connection scheme is called a \emph{cascade layer}. 

We can categorise the weights in a cascade network into four types: 
\begin{enumerate}
\item \textbf{Input to output layer weights} connecting the $d+1$ inputs to the
$q$ outputs, forming a perceptron network.
\item \textbf{Input to hidden layer weights} connecting the $d+1$ inputs to the $L$
 hidden cascade layers.
\item \textbf{Hidden to hidden layer weights} connecting the output of all the previous hidden cascade
layers $l_{1},l_{2},\ldots,l_{i-1}$, to the hidden cascade layer $l_{i}$.
\item \textbf{Hidden to output layer weights} connecting the outputs of the cascade
layers $l_{1},l_{2},\ldots,l_{L}$ to the $q$ output units.
\end{enumerate}

Figure \ref{fig:cascade_h} shows an example of a simple cascade neural network with three inputs, two output classes, and three hidden cascade layers ($l_{1}$, $l_{2}$, and $l_{3}$). All connections flow from left to right.  The cascade network is grown dynamically, one layer at a time, and the four different types of weights are each trained in slightly different ways (explained in detail below). Once training is complete, prediction uses a straight-forward feedforward algorithm that propagates values through the cascade layers.

\subsubsection{Training\label{subsec:Training}}

\begin{figure*}[!t]
\subfloat[Phase I, initial perceptron network.\label{fig:cascade_a}]{\includegraphics[width=0.22\textwidth]{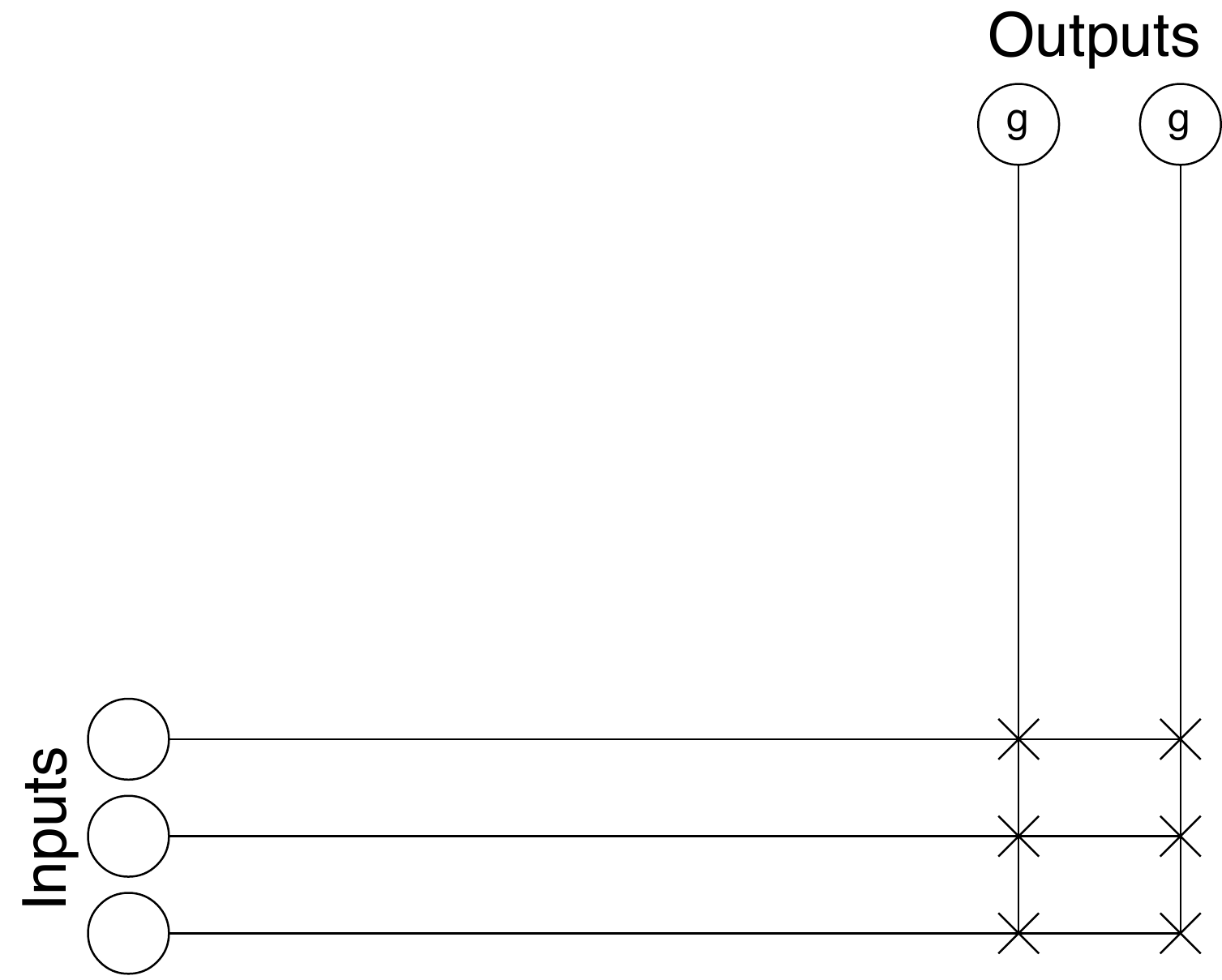}

}\,\,\,\,\,\,\,\,\subfloat[Phase II, train cascade layer 1 connections. Output layer weights
frozen.\label{fig:cascade_b}]{\includegraphics[width=0.22\textwidth]{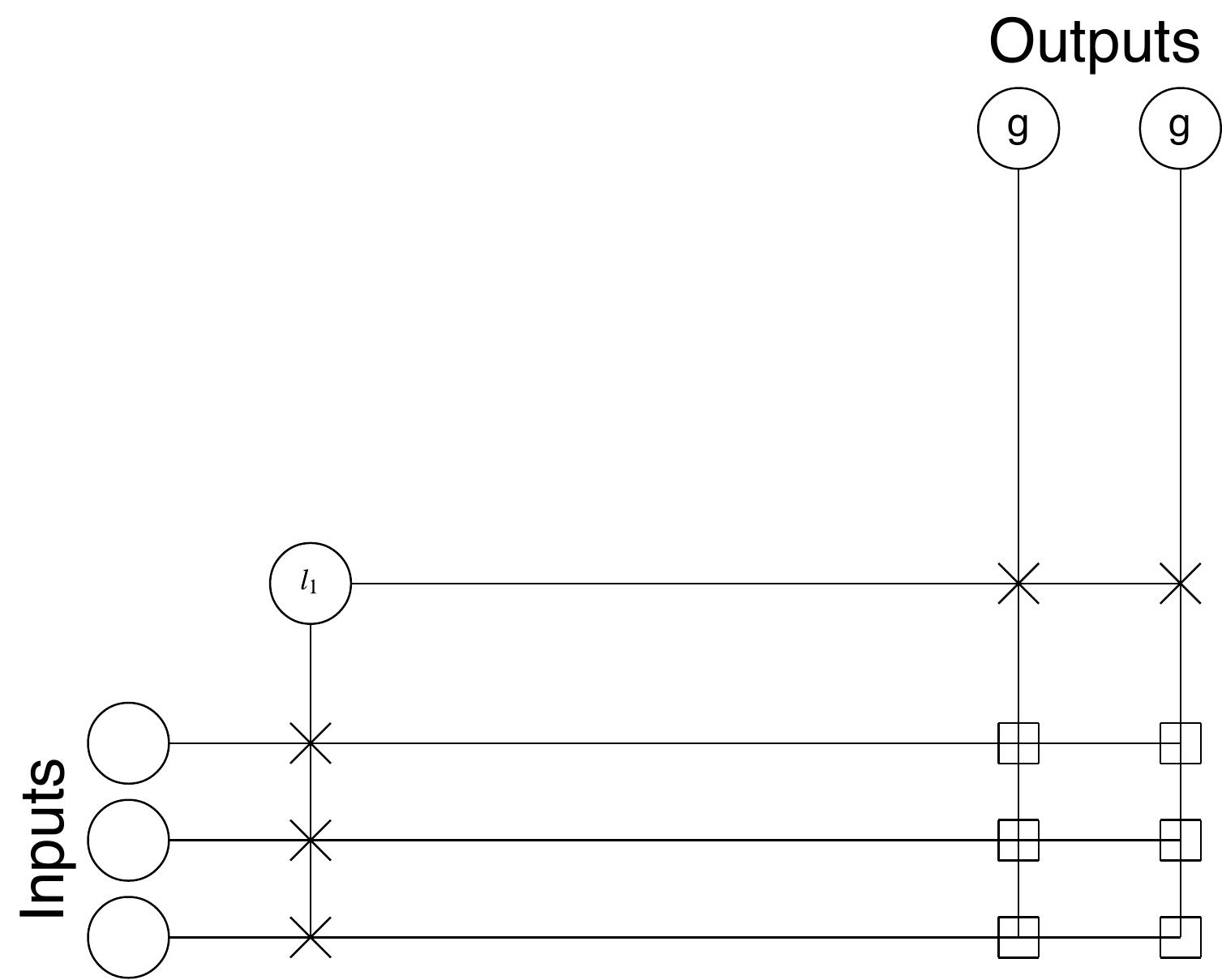}

}\,\,\,\,\,\,\,\,\subfloat[Phase I, cascade layer 1 added, training output weights, input to
hidden weights frozen.\label{fig:cascade_c}]{\includegraphics[width=0.22\textwidth]{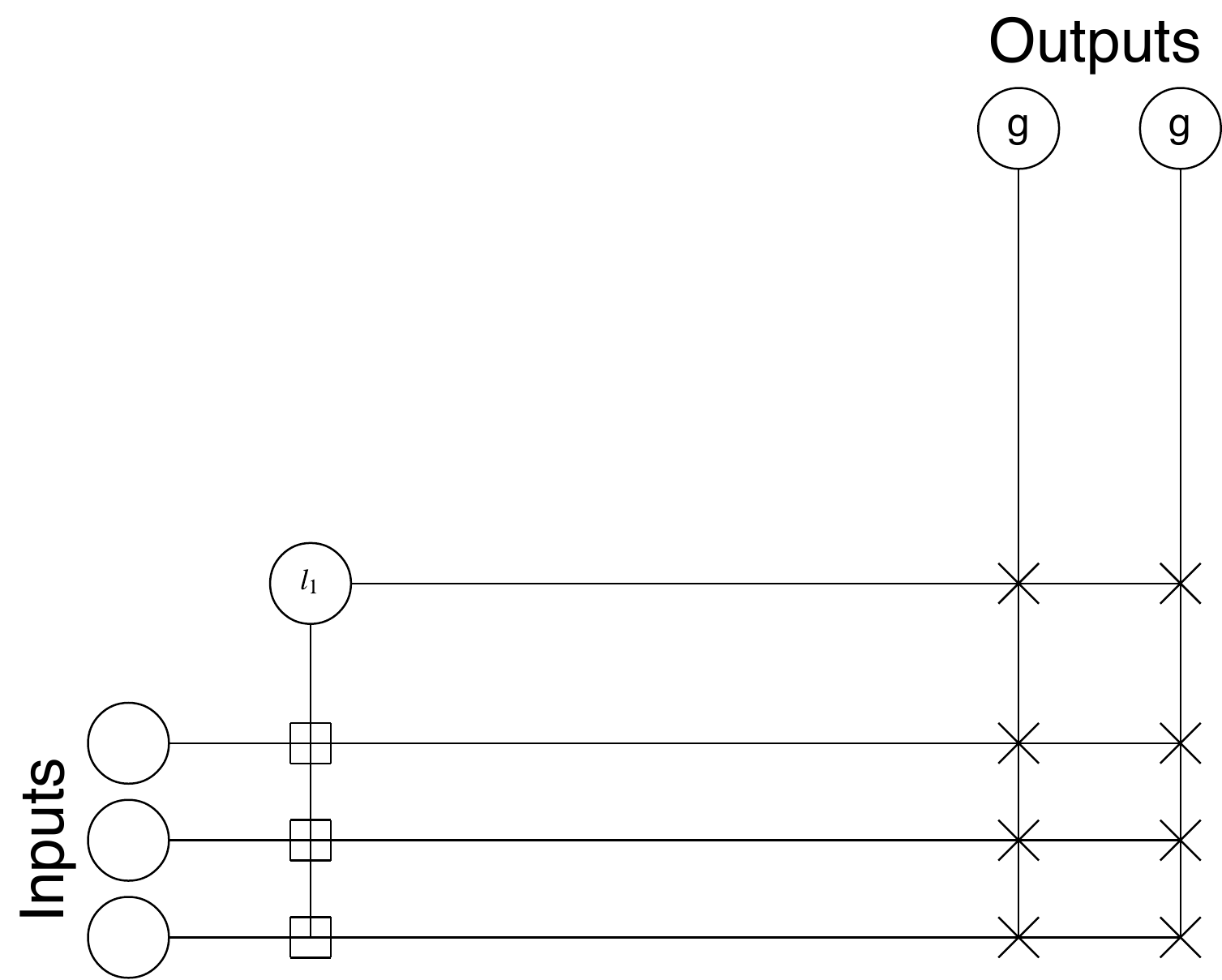}

}\,\,\,\,\,\,\,\,\subfloat[Phase II, train cascade layer 2 connections. Previous cascade layer
weights and output layer weights frozen.\label{fig:cascade_d}]{

\includegraphics[width=0.22\textwidth]{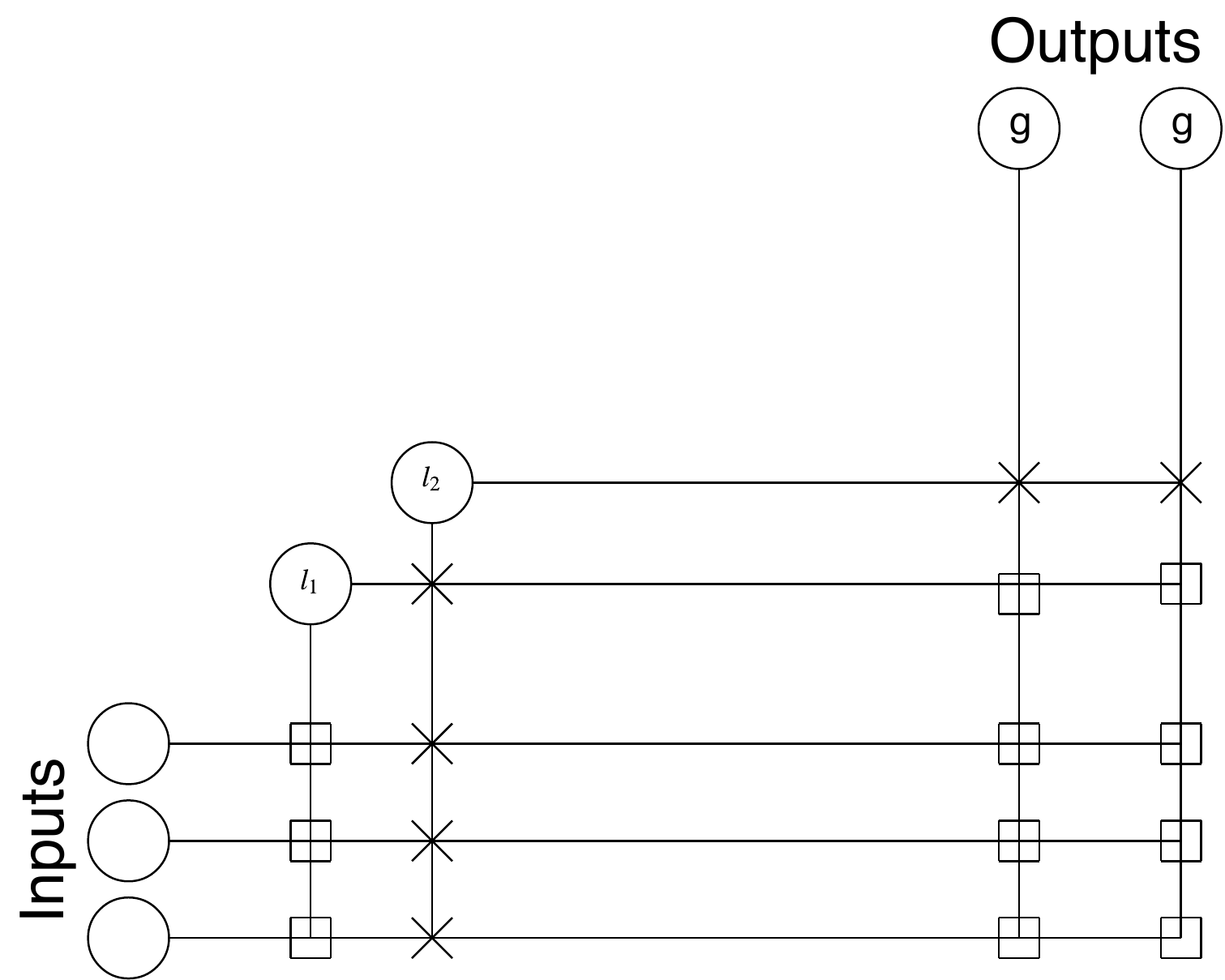}}\,\,\,\,\,\,\,\,

\subfloat[Phase I, cascade layer 2 added, training output weights, input to
hidden and hidden to hidden weights frozen.\label{fig:cascade_e}]{\includegraphics[width=0.22\textwidth]{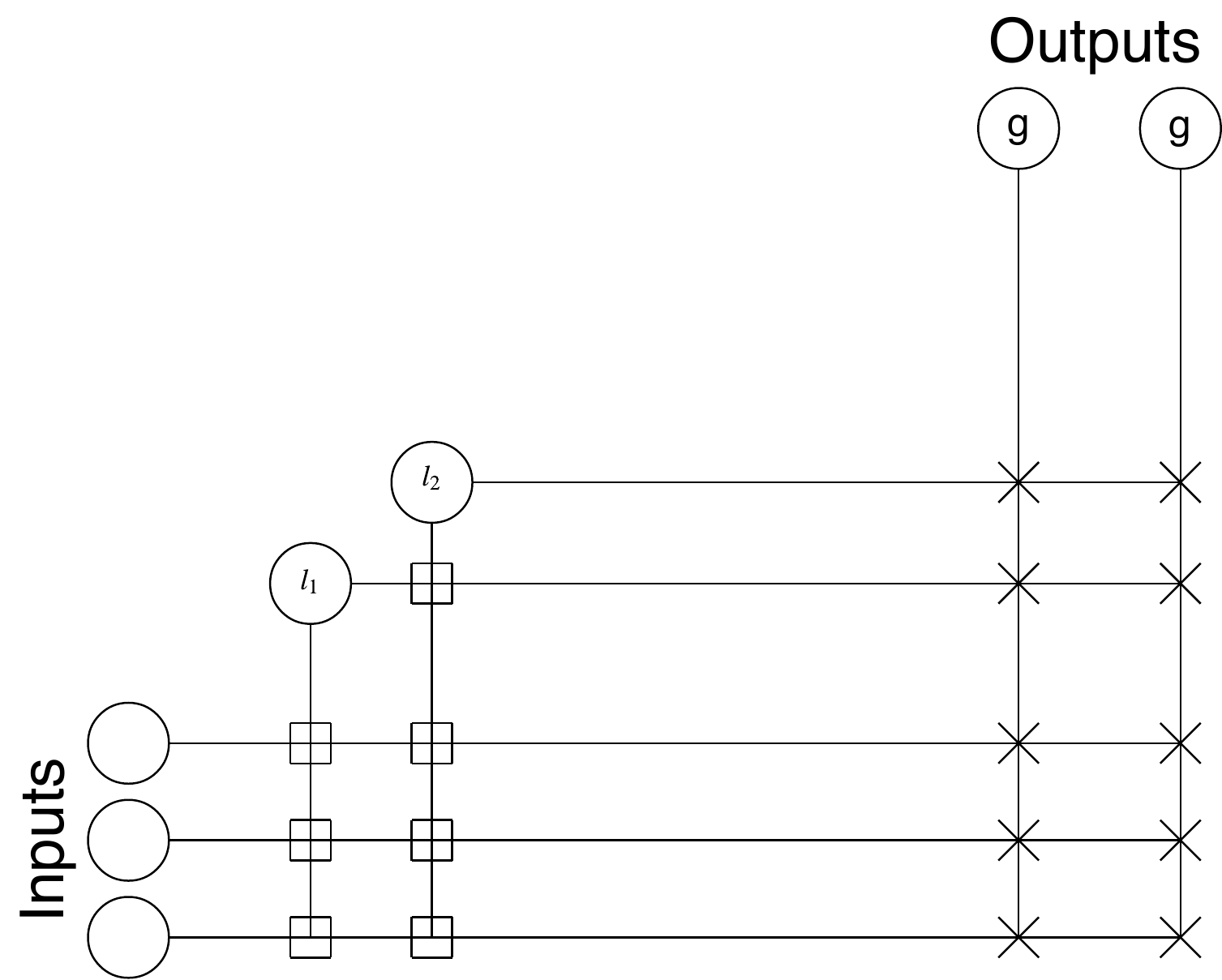}

}\,\,\,\,\,\,\,\,\subfloat[Phase II, train cascade layer 3 connections. Previous cascade layer
weights and output layer weights frozen.\label{fig:cascade_f}]{\includegraphics[width=0.22\textwidth]{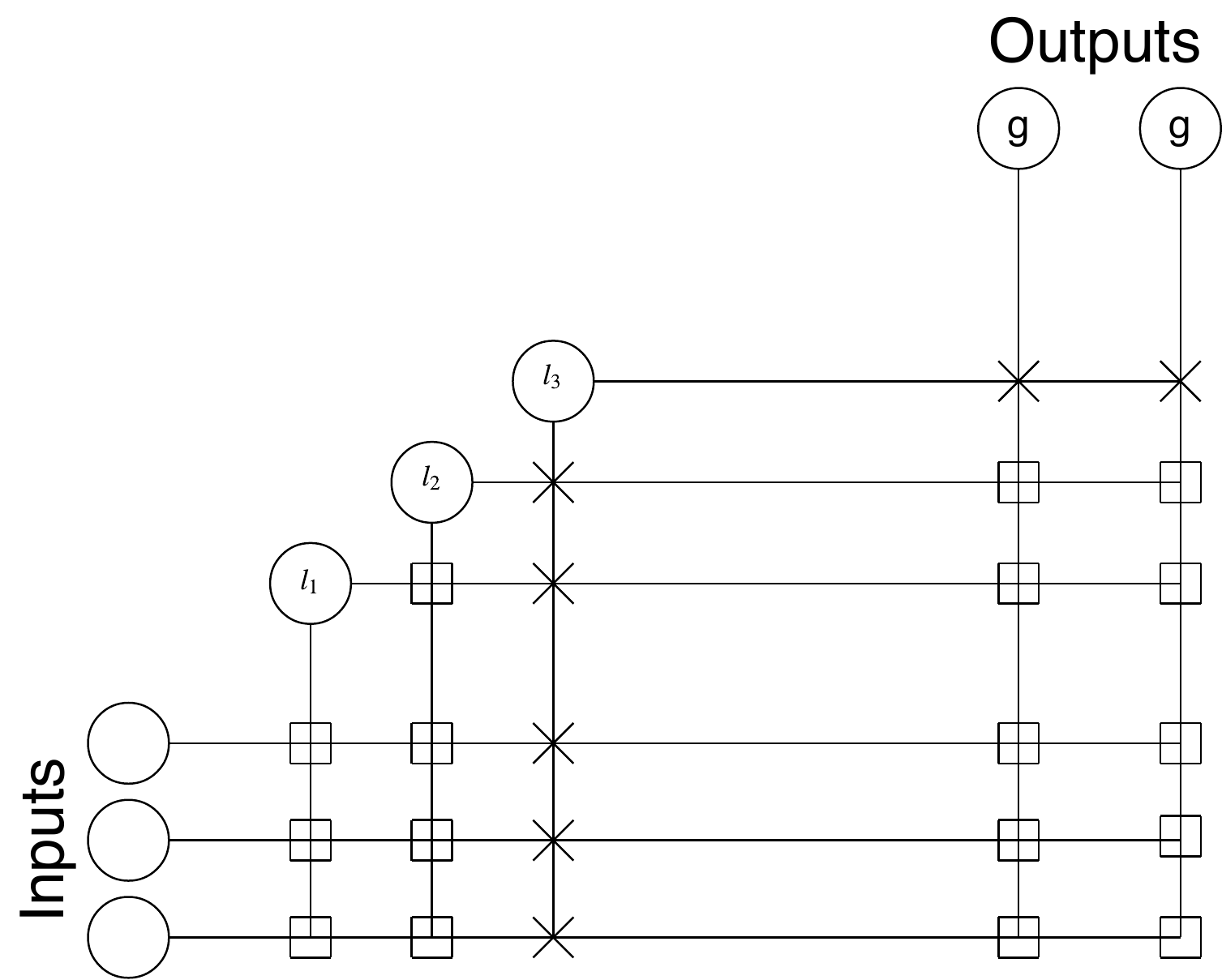}

}\,\,\,\,\,\,\,\,\subfloat[Phase I, cascade layer 3 added, training output weights, input to
hidden and hidden to hidden weights frozen.\label{fig:cascade_g}]{\includegraphics[width=0.22\textwidth]{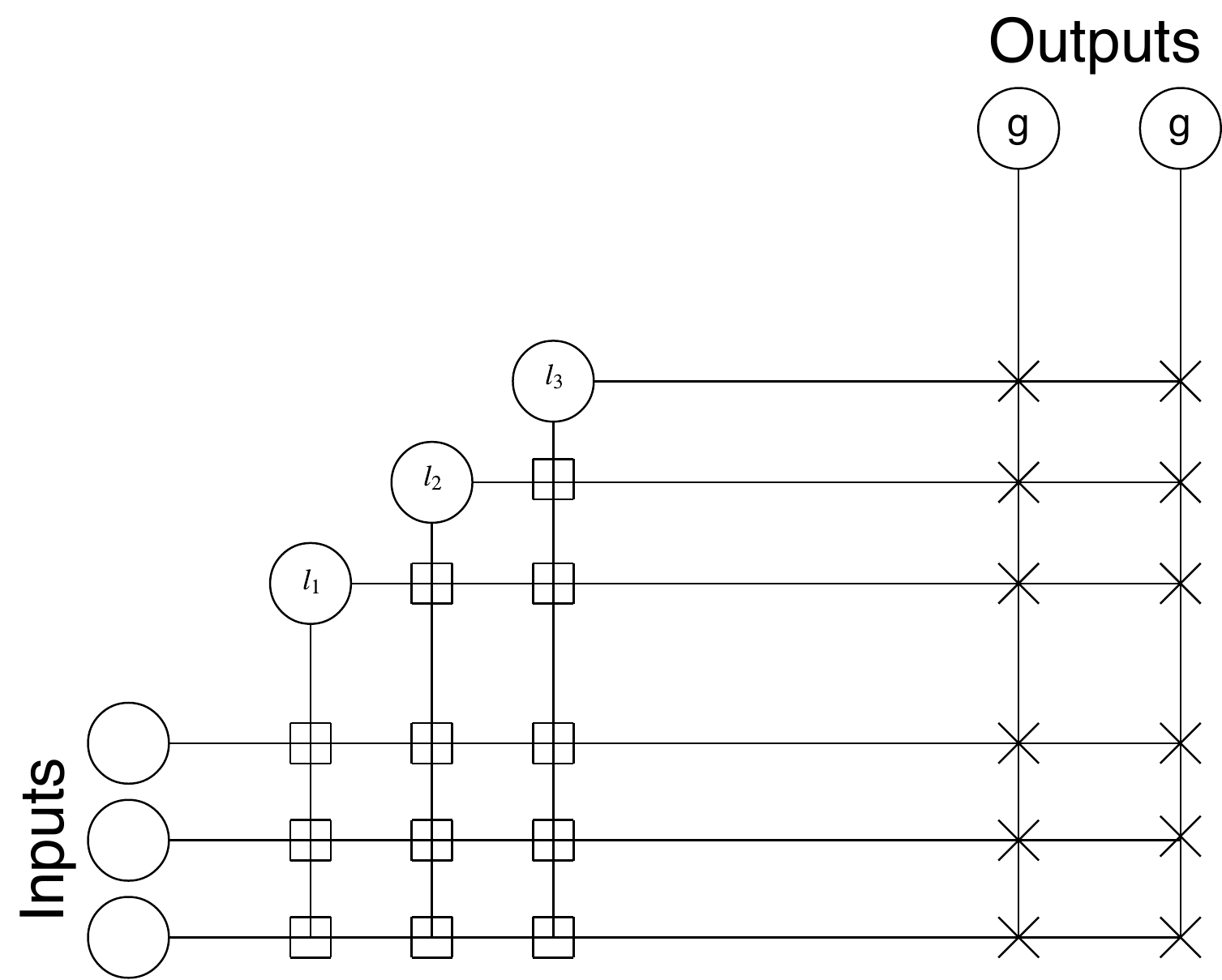}

}\,\,\,\,\,\,\,\,\subfloat[Final trained architecture, equivalent to Fig. \ref{fig:cascade_g}.\label{fig:cascade_h}]{\includegraphics[width=0.22\textwidth]{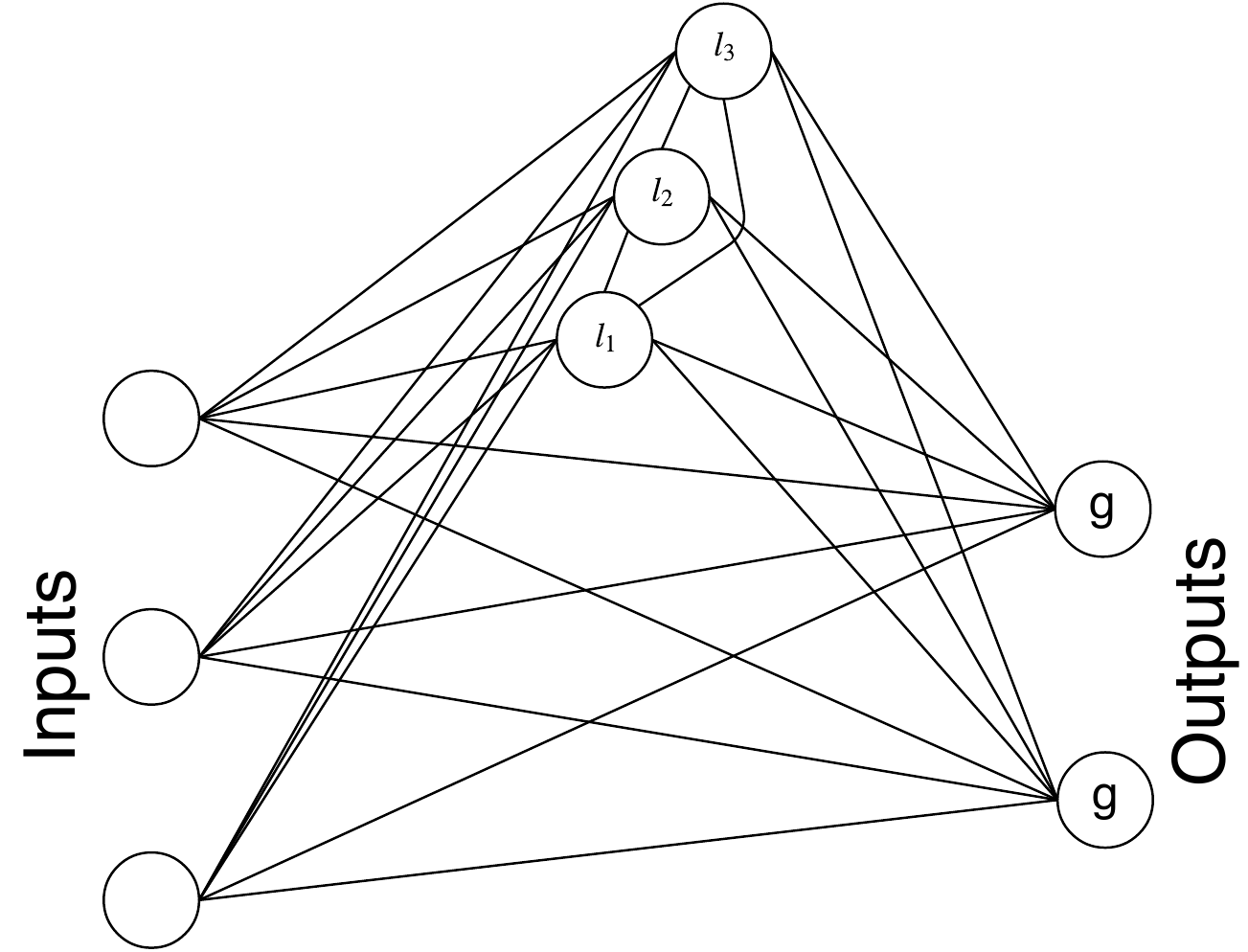}

}\,\,\,\,\,\,\,\,

\caption{The steps in the Cascade 2 network training algorithm. In each diagram the circles labelled \emph{Inputs} correspond to the input layer and the circles labelled \emph{Outputs} correspond to the output layer of the network. Hidden cascade units are represented by the circles labelled $l_{i}$. A weight between nodes in two layers exists, where horizontal and vertical lines intersect. Crosses indicate a weight that is trainable
at a specific step in the training process, while squares indicate a weight that is frozen at a specific step. $g$ is the activation
function used at the output layer. (h) shows a more typical network diagram of the final network trained.\label{fig:Cascaded-neural-network}}
\end{figure*}

Model training in the Cascade2 algorithm starts with a simple perceptron network (Figure \ref{fig:cascade_a})
with $d+1$ inputs and $q$ outputs. This network is referred to as the \emph{main network}. The main network is grown as training proceeds by iteratively adding hidden cascade layers to it. This is achieved by iteratively repeating two phases, \emph{Phase I} and \emph{Phase II}, each of which trains different parts of the cascade network. 

In Phase I, the input to output layer weights (type 1 in the list above) and hidden to output layer weights (type
4) of the main network are trained, while all other weights (input to hidden layer and hidden to hidden layer)
are frozen. The target values used in this phase to calculate the loss of the network are the target classes from the original dataset. The mean squared error (MSE) between the output of the main network and the ground truth is minimised using gradient descent.

Phase II trains and adds a new cascade layer $l_{i}$ at the $i^{th}$
iteration of training. The inputs to the newly added layer $l_{i}$ are
the $d+1$ input dimensions, and the outputs from the previous hidden
layers $l_{1},\dots,l_{i-1}$, in the main network. At this phase only the weights  involving the new hidden layer,  $l_{i}$, are trained. These are the input to hidden layer weights (type 2) for $l_{i}$; hidden to hidden layer weights on connections of the output of previous hidden cascade layers, $l_{1},\dots,l_{i-1}$, to the current hidden layer, $l_{i}$ (type 3); and the weights connecting the new hidden layer, $l_{i}$, to the output layer (type 4). All other weights in the main network are frozen. In this phase the target values used in training are not the original target values, but rather the error between the MSE of the main network constructed up to the previous iteration $i-1$, and the output of the new layer $l_{i}$.

Once the weights associated with the new hidden layer have been trained the layer and these weights are added to the main network. The weights connecting the new hidden layer, $l_{i}$, to the output layer are negated when these are added to the main network. This is so that the contribution of the output of the newly added layer will minimise the error of the main network \cite{nissen2007large}---recall that the newly added layer was trained to predict the main network error. 

When Phase II is complete, the algorithm proceeds again to Phase I and continues iterating between Phase I and Phase II until a maximum depth is reached 
or a learning error threshold is not exceeded. Training always ends with
Phase I.

\subsubsection{Example\label{subsec:Example}}

Figure \ref{fig:Cascaded-neural-network} shows an example of the
growth of a cascade network (the neural network diagram scheme used by Fahlman \& Lebiere \cite{NIPS1989_207}
is used). Figure \ref{fig:cascade_a} shows the
initial network with $3$ inputs and $2$ outputs. In this schematic
the intersections of the straight lines indicate the weights. A cross
at an intersection indicates that a weight is trainable at the current phase, while a square indicates that a weight is frozen. The algorithm starts in Phase
I and the network in Figure \ref{fig:cascade_a} is trained. All input to output layer weights (type 1), are trained (no hidden to output layer weights (type 4) exist yet). Next, in Phase II,
a new cascade layer, $l_{1}$, is added as shown in Figure \ref{fig:cascade_b},
and only the input to hidden layer (type 2) and hidden to hidden layer (type 3) weights related to the newly added
layer, $l_{1}$, are trained. Next, the process goes back to Phase I
and trains input to output layer (type 1) and hidden to output layer (type 4) weights in the main network as shown
in Figure \ref{fig:cascade_c}. This process iterates two more times
through Phase I and II in Figures \ref{fig:cascade_d}, \ref{fig:cascade_e} and
\ref{fig:cascade_f} until  the final network in Figure \ref{fig:cascade_g} is produced. Figure \ref{fig:cascade_h} shows this same final network using a more typical network diagram.

\section{The CascadeML Algorithm\label{sec:Proposed-Method:-CascadeML}}

CascadeML is a cascaded neural network approach to multi-label classification
based on Cascade2 \cite{PRECHELT1997885}.
The main objective of this method is to find good multi-label classifier
models that take advantage of label associations, while minimising the model selection and training time by omitting
hyperparameter tuning and architecture tuning.

CascadeML uses a similar training process to that described in Section
\ref{sec:Cascade-Correlation-Neural}. CascadeML starts with a perception
network with $d+1$ inputs (including the bias unit) and $q$ output
units, one for each label. In Phase I, only the hidden to output layer and
input to output layer weights are trained, as in Cascade2. The
loss function used in this phase is the BPMLL loss function shown in Eq. \eqref{eq:bpmll_cost}, which
allows CascadeML to consider pairwise label associations.

In Phase II CascadeML differs from Cascade2 in the following way. First, instead of adding hidden cascade layers with a single unit at each iteration, hidden cascade layers with multiple units are added. This gives rise to a hyperparameter selection problem as the number of units in each hidden layer needs to be determined. To overcome this, at each iteration of CascadeML, a \emph{candidate pool} of many candidate hidden cascade layers is trained, that could be added to the main network. Each of the candidate hidden cascade layers is initialised with randomly selected initial weight values, a randomly selected activation function, and a randomly selected number of units.
Each of the candidate hidden cascade layers is trained independently in parallel, to minimise MSE as explained in the Cascade2 algorithm. Once they have all been trained the best candidate hidden cascade layer from the candidate pool is selected (based on calculated loss on a validation dataset) and added to the main network.

To add flexibility to the network architectures explored by Cascade-ML, the algorithm can include candidate hidden cascade layers that are \emph{sibling} layers to the deepest hidden cascade layer already in the main network \cite{Baluja-1994-13783}, as well as \emph{successor} cascade layers. This allows wide architectures as well as deep architrectures to be explored. This is done by training candidate cascade networks in the candidate pool as successors and siblings and then selecting the best of the two types of candidate network.

The candidate hidden cascade layers in the candidate pool can each be trained independently in isolation from the main network, because when training the candidate hidden cascade layer, $l_{i}$, the inputs to the layer, the targets and the weights of the main network are all fixed. Therefore, the hidden cascade layer, $l_{i}$, can be considered a subnetwork, trained in isolation and then added to the main network.

When the best candidate hidden cascade layer is selected from the candidate pool, it is added to the main network by copying the input to hidden layer, weights to the main network, negating
the hidden to output layer weights and connecting them to the main network as in Cascade2. The main network increases in depth or the deepest
layer grows in breadth depending on whether a successor or a sibling candidate layer was selected.

For both Phase I and Phase II, an adaptive first order gradient descent algorithm iRProp- \cite{igel2000improving},
a variant of RProp \cite{igel2000improving,Rojas:1996:NNS:235222},
is used. iRProp- was found to be more stable than the originally used Quickprop \cite{Fahlman88anempirical}. iRProp- is an adaptive algorithm which uses an adaptive learning
rate and the sign of the partial derivative of the error function
for each weight adjustment. This method mainly helps learn very fast in the flat regions of the error space 
and near local minima, as it uses only the sign of the partial derivative (ignoring its magnitude) and uses an adaptive learning
rate. L2 regularisation \cite{Goodfellow-et-al-2016} was used in both phases of CascadeML.



\section{Experiment Design\label{sec:Experiment}}

To evaluate the effectiveness of CascadeML, an experiment was performed on ten well-known multi-label
benchmark datasets listed in Table \ref{tab:datasets}. In Table \ref{tab:datasets} \emph{Instances}, \emph{Inputs} and \emph{Labels} are the number of datapoints, the dimension of the datapoint and the number of labels respectively. \emph{Labelsets} indicates the number of unique combinations of labels present in a dataset. \emph{Cardinality} measures the average number of labels assigned to each datapoint, and \emph{MeanIR} \cite{scumble} indicates the imbalance ratio of the labels.
\begin{table}[!t]
\caption{Multi-label datasets\label{tab:datasets}}

\centering\centering
\setlength{\tabcolsep}{3pt}
\begin{tabular}{lrrrHrrr}
\hline 
Dataset  & Instances  & Inputs  & Labels  & Labelsets  & Labelsets  & Cardinality & MeanIR \tabularnewline
\hline 
flags  & 194 & 26 & 7 & 54 & 24 & 3.392 & 2.255\tabularnewline
yeast  & 2417  & 103  & 14  & 198  & 77  & 4.237  & 7.197\tabularnewline
scene  & 2407  & 294  & 6  & 15  & 3  & 1.074  & 1.254\tabularnewline
emotions  & 593  & 72  & 6  & 27  & 4  & 1.869  & 1.478\tabularnewline
medical  & 978  & 1449  & 45  & 94  & 33  & 1.245  & 89.501\tabularnewline
enron  & 1702  & 1001  & 53  & 753  & 573  & 3.378  & 73.953\tabularnewline
birds  & 322  & 260  & 20  & 89  & 55  & 1.503  & 13.004\tabularnewline
genbase  & 662  & 1186  & 27  & 32  & 10  & 1.252  & 37.315\tabularnewline
cal500  & 502  & 68  & 174  & 502  & 502  & 26.044  & 20.578\tabularnewline
llog  & 1460  & 1004  & 75  & 304  & 189  & 1.180  & 39.267\tabularnewline
\hline 
\end{tabular}
\end{table}

 
The performance of models trained using CascadeML was compared with the multi-label neural network
algorithm BPMLL, and four other state-of-the-art multi-label classification algorithms:
\emph{classifier chains} \cite{Read2011} (CC), \emph{RAkEL}
\cite{Tsoumakas2007RandomK}, HOMER \cite{tsoumakas2008effective}, and MLkNN \cite{mlknn}. These algorithms were selected to cover different types of multi-lable classification techniques. Classifier chains, RAkEL and HOMER, when used with SVMs, are ensemble methods that have been previously shown to be the best performing the multi-label classifiers \cite{pakrashi2016benchmarking,MADJAROV20123084}; BPMLL is a well-known multi-label specific neural network algorithm; and MLkNN is a nearest-neighbour based algorithm adaptation method. The implementations of classifier chains, RAkEL, HOMER, MLkNN and BPMLL are from the MULAN library \cite{mulan} and implemented in Java. CascadeML was implemented in Python\footnote{A version of CascadeML is available at: https://github.com/phoxis/CascadeML}.


To compare the performances of the methods, label-based macro-averaged F-Score \cite{zhang2014review} was used. This is preferred over Hamming loss \cite{zhang2014review}, used in several previous studies (e.g. \cite{brknn,mlknn,iblrml}), as when used with highly imbalanced multi-label datasets Hamming loss tends to allow the performance on the majority labels to overwhelm performance on the minority labels. Label-based macro-averaged F-Score does not suffer from this problem. For every dataset performance is evaluated using a 2 times 5-fold cross validation experiment. The mean label-based Macro-averaged F-Score from these experiments are reported.


%
%


\subsection{Configuring CascadeML}
Although there is no hyperparameter tuning required for CascadeML, it does require some configuration. In the experiments described here, at each iteration, the candidate pool contained two candidates for each combination of layer type---$successor$ or $sibling$---and activation unit type---$linear$, $sigmoid$, or $tanh$. This made for 12 candidate hidden cascade layers at each iteration. The number of hidden units in each candidate layer was selected randomly selected as a fraction of the number of input dimensions, $d$, following a uniform distribution  in $(0,1]$.

For the output layer of the main network the activation function used was $tanh$ as the cost function requires bipolar encoding of the labels. During Phase II of training the outputs of the candidate layers in the pool use a linear activation function, as explained in Section \ref{sec:Proposed-Method:-CascadeML}. L2 regularisation was used in all training phases with regularisation value of $10^{-5}$. In Phase I and Phase II training early stopping is used where training stops if the average loss (based on a validation dataset) calculated over a  window of the last $20$ training epochs increases from one iteration to the next. For both Phase I and Phase II iRProp- is initialised as recommended in \cite{igel2000improving}. The maximum number of iterations allowed for iRProp- in both phases was $2000$. To set an upper bound on network growth in CascadeML two stopping criteria were used: (1) a new cascade layer (sibling or successor) was added only if did not lead to an increase in the validation loss of the entire network, and (2) only $20$ iterations are allowed. 

\subsection{Configuring other algorithms}
All the algorithms used in the experiment, except CascadeML,
underwent a grid search based hyperparameter tuning using 2 time 5-folds cross validation. For classifier chains, RAkEL and HOMER, support vector machines \cite{Kelleher:2015:FML:2815672}
with a radial basis kernel (SVM-RBF) were used as the base classifier. In these cases 12 combinations of the regularisation parameter, $C$, and the kernel spread, $\sigma$, were included the hyperparameter grid. For RAkEL the subset size hyperparameter (ranging from 2 to 6) was also included, and for HOMER the cluster size hyperparameter (ranging from 2 to 6) was also included. 
For BPMLL the only hyperparameter in the grid search was the number of units in the hidden layer. Sizes of  20\%, 40\%, 60\%, 80\% and 100\% of the number of inputs for each dataset were explored, as recommended by
Zhang et al. \cite{Zhang:2006:MNN:1159162.1159294}. In this case the L2 regularisation coefficient was set to $10^{-5}$ and a maximum of $10000$ iterations were allowed, based on \cite{pakrashi2016benchmarking}.

The results presented are based on the best performing hyperparameter combinations. Finally, the mean label-based Macro-averages F-Scores of 2 times 5-folds cross validation experiments of the best hyperparameter combination are reported.

\section{Results\label{sec:Results}}

The results of the experiments are shown in Table \ref{tab:Multi-label-Macro-averaged},
where the columns indicate the algorithms and the rows indicate the
datasets. Each cell of the table shows the label-based macro-averaged
F-Score (higher values are better) followed by the standard deviation over the cross valition folds. These label-based
F-Scores are computed through extensive cross validated hyperparameter
tuning. The values in the parenthesis indicate
the relative ranking (lower values are better) of the algorithm with
respect to the corresponding dataset. The last row of Table \ref{tab:Multi-label-Macro-averaged}
indicates the overall average ranks of the corresponding algorithms.

\begin{table}[!t]
\caption{Results of experiments. Rows indicate the datasets, columns indicate
algorithms. Values in cells are mean label-based macro-averaged F-Scores and the standard deviations
followed by relative ranks in parenthesis. Last row are the average
ranks of the corresponding algorithms.\label{tab:Multi-label-Macro-averaged}}

\centering\centering
\resizebox{\textwidth}{!}{%
\setlength{\tabcolsep}{5pt}
\begin{tabular}{lllllll}
  \hline
          & CascadeML           & RAkEL               & CC                  & BPMLL               & HOMER               & MLkNN                \\ 
  \hline
  flags   & 0.6723$\pm$0.06 (1) & 0.6505$\pm$0.04 (2) & 0.6405$\pm$0.06 (4) & 0.5948$\pm$0.03 (6) & 0.6479$\pm$0.04 (3) & 0.6009$\pm$0.07 (5)  \\ 
  yeast   & 0.4624$\pm$0.01 (1) & 0.4367$\pm$0.02 (4) & 0.4510$\pm$0.01 (2) & 0.4357$\pm$0.01 (5) & 0.4478$\pm$0.02 (3) & 0.3772$\pm$0.01 (6)  \\ 
  scene   & 0.7606$\pm$0.01 (5) & 0.8017$\pm$0.01 (2) & 0.8040$\pm$0.01 (1) & 0.7777$\pm$0.01 (4) & 0.8001$\pm$0.02 (3) & 0.7424$\pm$0.02 (6)  \\ 
  emotions& 0.6671$\pm$0.02 (2) & 0.6281$\pm$0.02 (4) & 0.6242$\pm$0.01 (5) & 0.6899$\pm$0.02 (1) & 0.6212$\pm$0.02 (6) & 0.6294$\pm$0.03 (3)  \\ 
  medical & 0.6758$\pm$0.02 (3) & 0.6966$\pm$0.03 (1) & 0.6924$\pm$0.04 (2) & 0.5582$\pm$0.08 (5) & 0.6108$\pm$0.05 (4) & 0.5398$\pm$0.05 (6)  \\ 
  enron   & 0.2852$\pm$0.02 (3) & 0.2882$\pm$0.04 (2) & 0.2890$\pm$0.03 (1) & 0.2806$\pm$0.02 (5) & 0.2812$\pm$0.03 (4) & 0.1771$\pm$0.03 (6)  \\ 
  birds   & 0.4812$\pm$0.03 (1) & 0.1812$\pm$0.06 (4) & 0.1582$\pm$0.06 (5) & 0.3426$\pm$0.06 (2) & 0.1551$\pm$0.05 (6) & 0.2256$\pm$0.09 (3)  \\ 
  genbase & 0.9403$\pm$0.02 (3) & 0.9432$\pm$0.05 (2) & 0.9440$\pm$0.04 (1) & 0.8149$\pm$0.12 (6) & 0.9394$\pm$0.05 (4) & 0.8502$\pm$0.05 (5)  \\ 
  cal500  & 0.2263$\pm$0.01 (2) & 0.1790$\pm$0.01 (5) & 0.1849$\pm$0.01 (4) & 0.2367$\pm$0.02 (1) & 0.1988$\pm$0.02 (3) & 0.1007$\pm$0.01 (6)  \\ 
  llog    & 0.2264$\pm$0.03 (6) & 0.2998$\pm$0.05 (1) & 0.2916$\pm$0.03 (3) & 0.2953$\pm$0.06 (2) & 0.2561$\pm$0.03 (5) & 0.2630$\pm$0.05 (4)  \\ 
  \hline
  Avg. rank & \multicolumn{1}{r}{2.7\ } & \multicolumn{1}{r}{2.7\ } & \multicolumn{1}{r}{2.8\ } & \multicolumn{1}{r}{3.7\ } & \multicolumn{1}{r}{4.1\ } & \multicolumn{1}{r}{5.0\ } \tabularnewline
  \hline 
\end{tabular}}
\end{table}

Table \ref{tab:Multi-label-Macro-averaged} shows that CascadeML (avg.
rank 2.7) performed better than BPMLL (avg. rank 3.7), HOMER (avg. rank 4.1), 
MLkNN (avg. rank 5.0) and  CC (avg. rank 2.8), overall. RAkEL had the same overall average rank as CascadeML. 

Although RAkEL and CC had similar rank as CascadeML
on average, it must be noted that the label-based macro-averaged
F-Score for CC as well as for the other methods were achieved after doing
an extensive hyperparameter tuning which CascadeML did not require. Besides tuning for C and sigma hyperparameter of the underlying SVM-RBFs for RAkEL, CC and HOMER, there are other hyperparameters which needs to be tuned. For RAkEL the labelset size needs to be selected, for CC the chain order needs to be defined, and for HOMER the clustering algorithma and the cluster size needs to be defined. All of these hyperparameter increases the hyperparameter search space dimensionality.

Absolute running times or the number of operations are not directly comparable as the methods are different from CascadeML and implementations of the algorithms span different programming languages. However, it is worth noting that the completion of the CC, RAkEL benchmarks took multiple weeks (with multiple folds run in parallel) due to the hyperparameter tuning involved, whereas running the equivalent benchmark for CascadeML took less than a week.

The nature of the incremental growth and training in combination with the fast convergence nature of iRProp- with the L2 regularisation helped the network to generalise as well as converge faster. Also, note that the candidate unit pool size was set to 12 and all of them were run in parallel, hence the real runtime of the candidate training would be the maximum time taken of the 12 of the candidates. Therefore, by exploiting the cascade architecture and training process, as well as using the iRProp- algorithm along with L2 regularisation, CascadeML was able to maintain a very good level of performance without hyperparameter tuning.

Figure \ref{fig:cascade-depth-d} shows the training costs for scene dataset for one fold. The vertical dotted line indicates the addition of a candidate layer to the main network. After each addition of the candidate network the cost increases but then sharply decreases at first then continues decreasing steadily.

%
%

\begin{figure*}[t!]
\subfloat[yeast dataset]{
\includegraphics[width=0.5\textwidth]{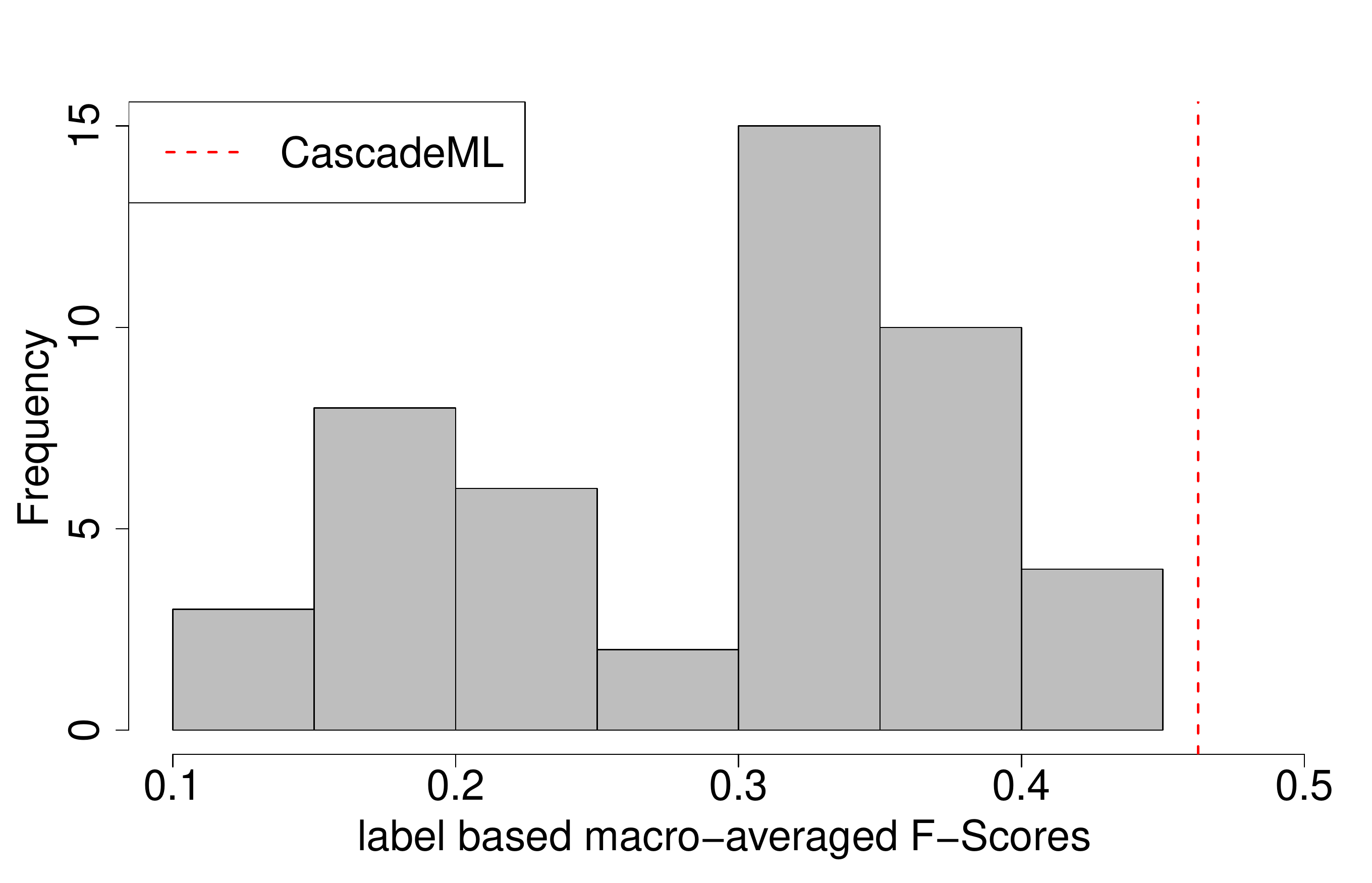}}\subfloat[enron dataset]{

\includegraphics[width=0.5\textwidth]{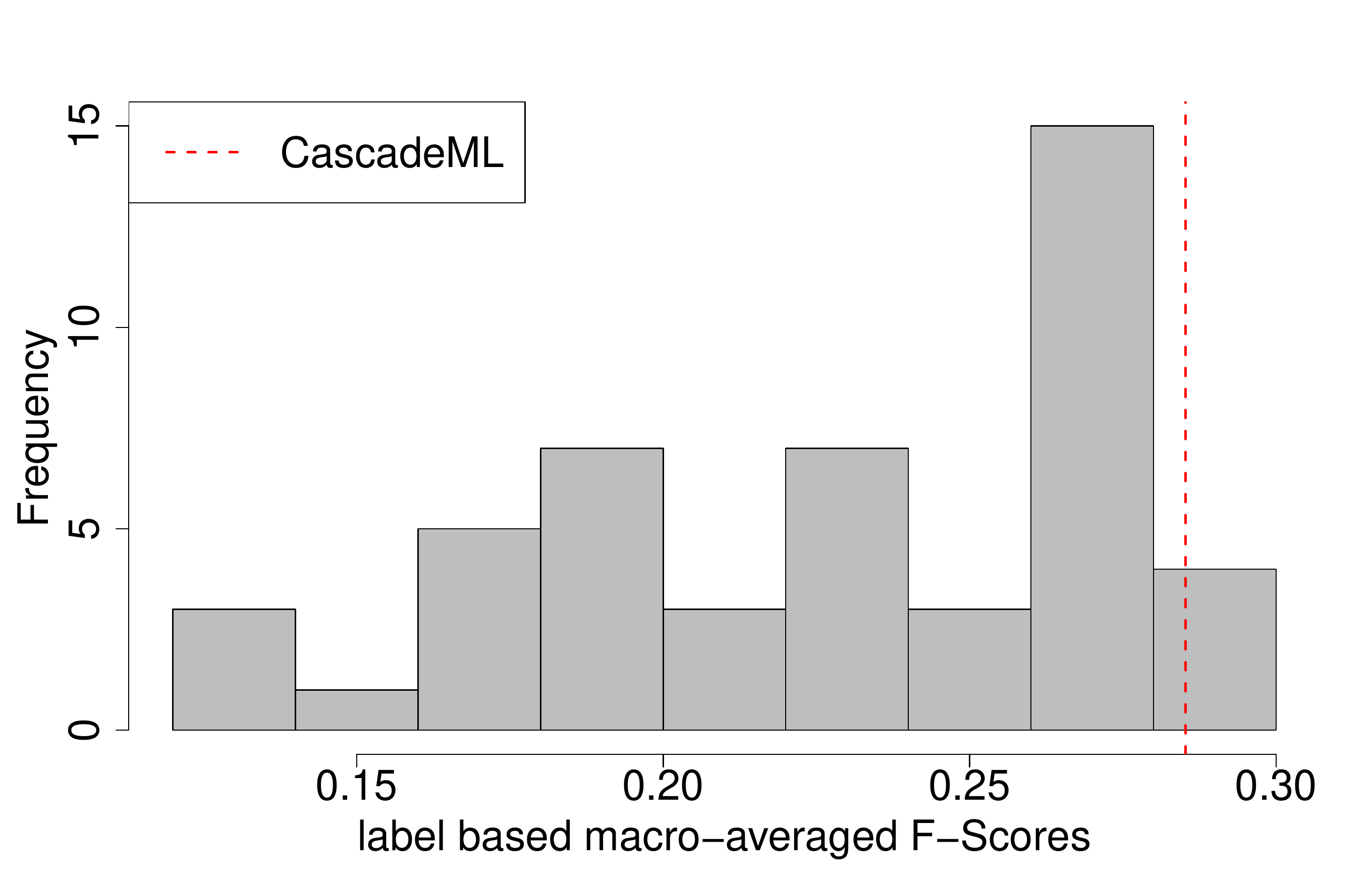}}

\caption{Histogram of label-based macro-averaged F-Scores achieved from all
hyperparameter combinations of subset size for RAkEL and C, sigma for the underlying SVM-RBFs. The vertical dotted line indicates CascadeML's performance.\label{fig:Classifier-chain-hist}}
\end{figure*}

%
%
%

\begin{figure*}[t!]
\subfloat[Cascaded depth]{
\label{fig:cascade-depth-a}
\includegraphics[width=0.5\textwidth]{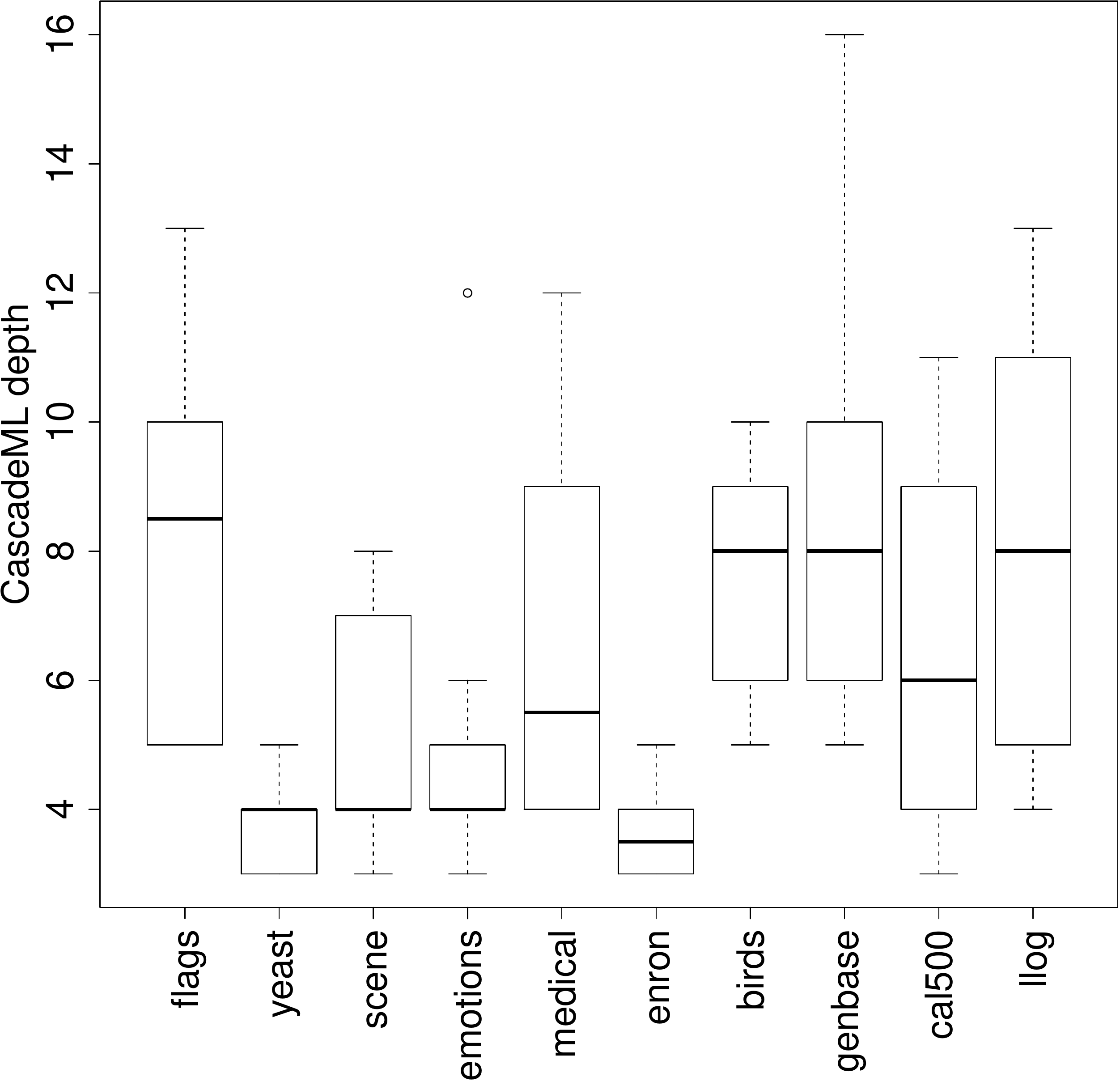}}\subfloat[Total hidden nodes scaled by input size\label{fig:cascade-depth-b}]{
\includegraphics[width=0.5\textwidth]{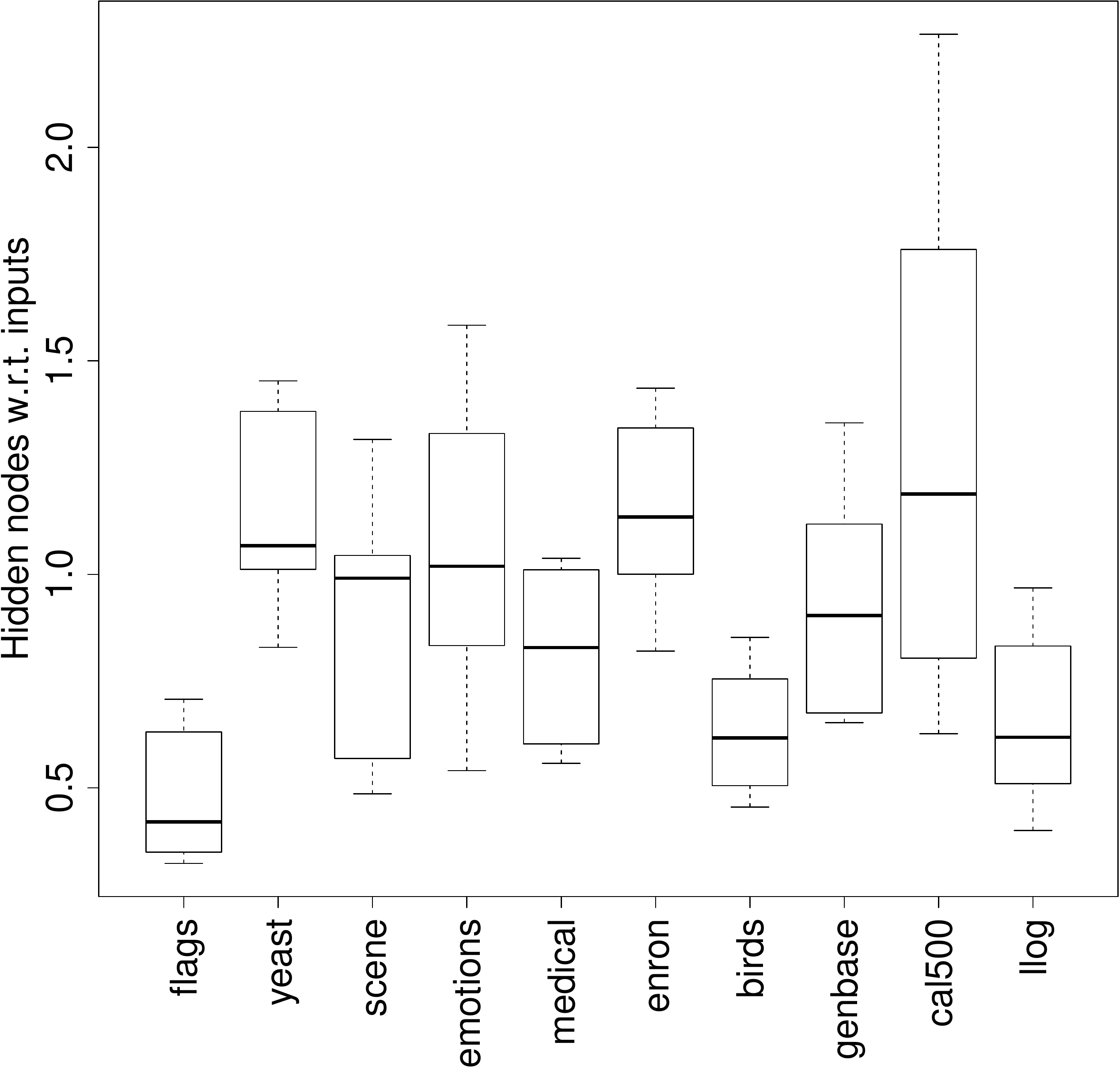}}

\subfloat[All datasets, all folds depth vs. scaled numbmer of hidden nodes\label{fig:cascade-depth-c}]{
\label{fig:cascade-depth-c}
\includegraphics[width=0.5\textwidth]{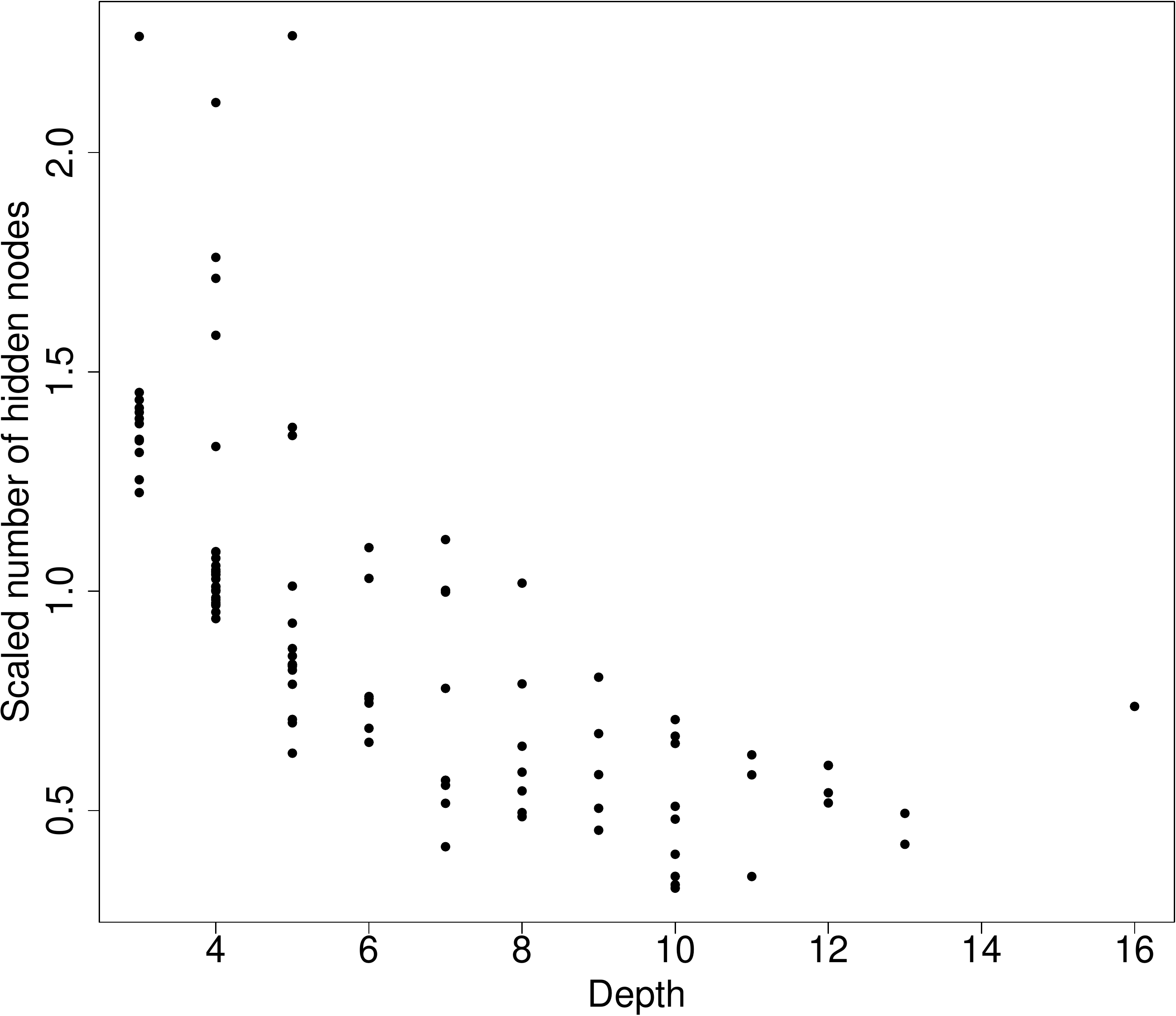}}\subfloat[Training costs for a fold for scene\label{fig:cascade-depth-d}]{
\label{fig:cascade-depth-d}
\includegraphics[width=0.5\textwidth]{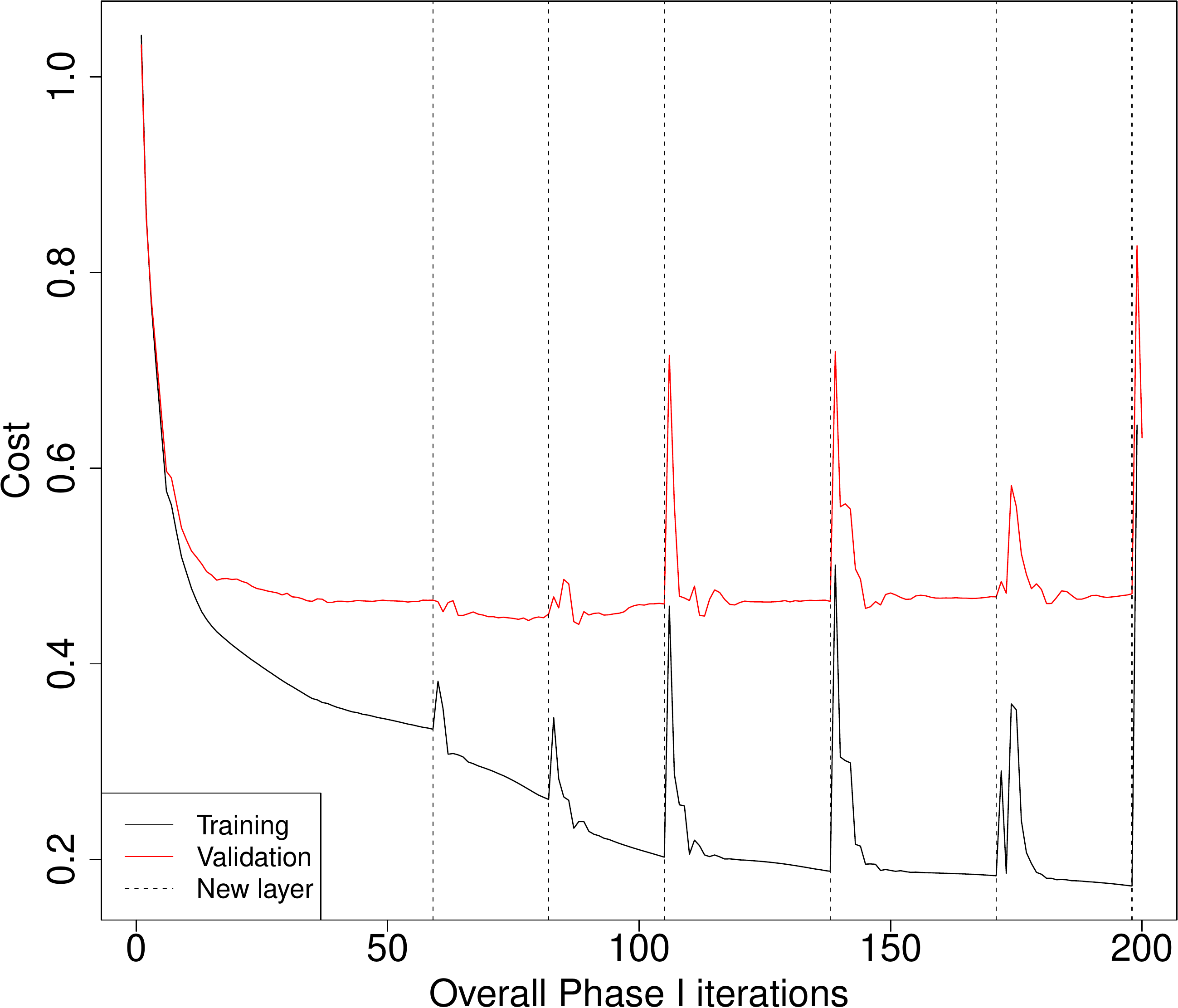}}

\caption{CascadeML trained network properties.\label{fig:cascade-depth}}
\end{figure*}

\begin{table}
\caption{Summary of the trained CascadeML network architecture for all datasets.\label{tab:cascademl-summary-arch}}

\centering
\resizebox{0.5\textwidth}{!}{%
\setlength{\tabcolsep}{15pt}
\begin{tabular}{lllH}
\hline 
          & Cascade         & Scaled hidden       & Normalised     \tabularnewline
          & Depth           & nodes               & nodes          \tabularnewline
\hline 
flags     & 8.30 $\pm$ 2.87 & 0.47 $\pm$ 0.15 & 0.67 $\pm$ 1.53  \tabularnewline
yeast     & 3.80 $\pm$ 0.79 & 1.16 $\pm$ 0.22 & 2.06 $\pm$ 3.22  \tabularnewline
scene     & 5.10 $\pm$ 1.85 & 0.89 $\pm$ 0.28 & 1.34 $\pm$ 2.90  \tabularnewline
emotions  & 5.00 $\pm$ 2.58 & 1.04 $\pm$ 0.32 & 1.61 $\pm$ 3.01  \tabularnewline
medical   & 6.80 $\pm$ 3.16 & 0.80 $\pm$ 0.20 & 1.03 $\pm$ 1.38  \tabularnewline
enron     & 3.60 $\pm$ 0.70 & 1.16 $\pm$ 0.21 & 2.18 $\pm$ 3.09  \tabularnewline
birds     & 7.60 $\pm$ 1.58 & 0.64 $\pm$ 0.14 & 0.65 $\pm$ 1.55  \tabularnewline
genbase   & 8.40 $\pm$ 3.24 & 1.04 $\pm$ 0.49 & 0.72 $\pm$ 1.31  \tabularnewline
cal500    & 6.40 $\pm$ 2.84 & 1.34 $\pm$ 0.60 & 1.29 $\pm$ 1.00  \tabularnewline
llog      & 8.10 $\pm$ 3.45 & 0.67 $\pm$ 0.20 & 0.78 $\pm$ 1.58  \tabularnewline
\hline 
\end{tabular}}
\end{table}

It is important to note that CascadeML has the advantage of not requiring hyperparameter tuning. For other algorithms the selection of hyperparameter values can have a huge impact on performance. For example, Figure \ref{fig:Classifier-chain-hist} shows the distribution of the label-based macro-averaged F-Scores for different combinations of label subset size, C and sigma values the underlying SVM-RBFs
of RAkEL for the yeast and enron datasets. Note that the F-Score values in Figure \ref{fig:Classifier-chain-hist} vary significantly. For the yeast dataset, CascadeML performed the best, and for enron dataset only 2.1 \% of the hyperparameter attained better result than CascadeML. In general the distribution skews towards models with relatively poorer performance. CascadeML attained similar high values of performance in both the cases ($0.4624$ for the yeast dataset and $0.2852$ for the enron datset) without any need for hyperparameter tuning.

Table \ref{tab:Multi-label-Macro-averaged} shows that, CascadeML is very competitive across different datasets compared to state-of-the-art algorithms while not requiring hyperparameter tuning.

CascadeML learns architectures with different number of nodes and activation functions per layer as shown in Table \ref{tab:cascademl-summary-arch} and Figure \ref{fig:cascade-depth} for every dataset over multiple folds. In Table \ref{tab:cascademl-summary-arch} \emph{Cascade depth} indicates the depth of the cascade network, and \emph{Scaled hidden nodes} indicate the number of total hidden units divided by the number of input units for each dataset. Figure \ref{fig:cascade-depth-a} shows the boxplots of the learnet network depths over folds for each dataset and Figure \ref{fig:cascade-depth-b} shows the boxplots for the scaled hidden nodes. Note that although the standard deviations of the performances in Table \ref{tab:Multi-label-Macro-averaged} are small, the trained layer depth and the scaled hidden nodes have high standard deviations. Figure \ref{fig:cascade-depth-c} shows a scatterplot of the depths and the scaled hidden nodes values over all datasets and folds. This indicates that the learned networks were either deep with fewer nodes per layer, or shallow but more nodes per layer, therefore having a similar network capacity and hence the F-Score performance over the folds were similar, although the architecture learned were very different.

\begin{figure*}[!t]

\includegraphics[width=\textwidth]{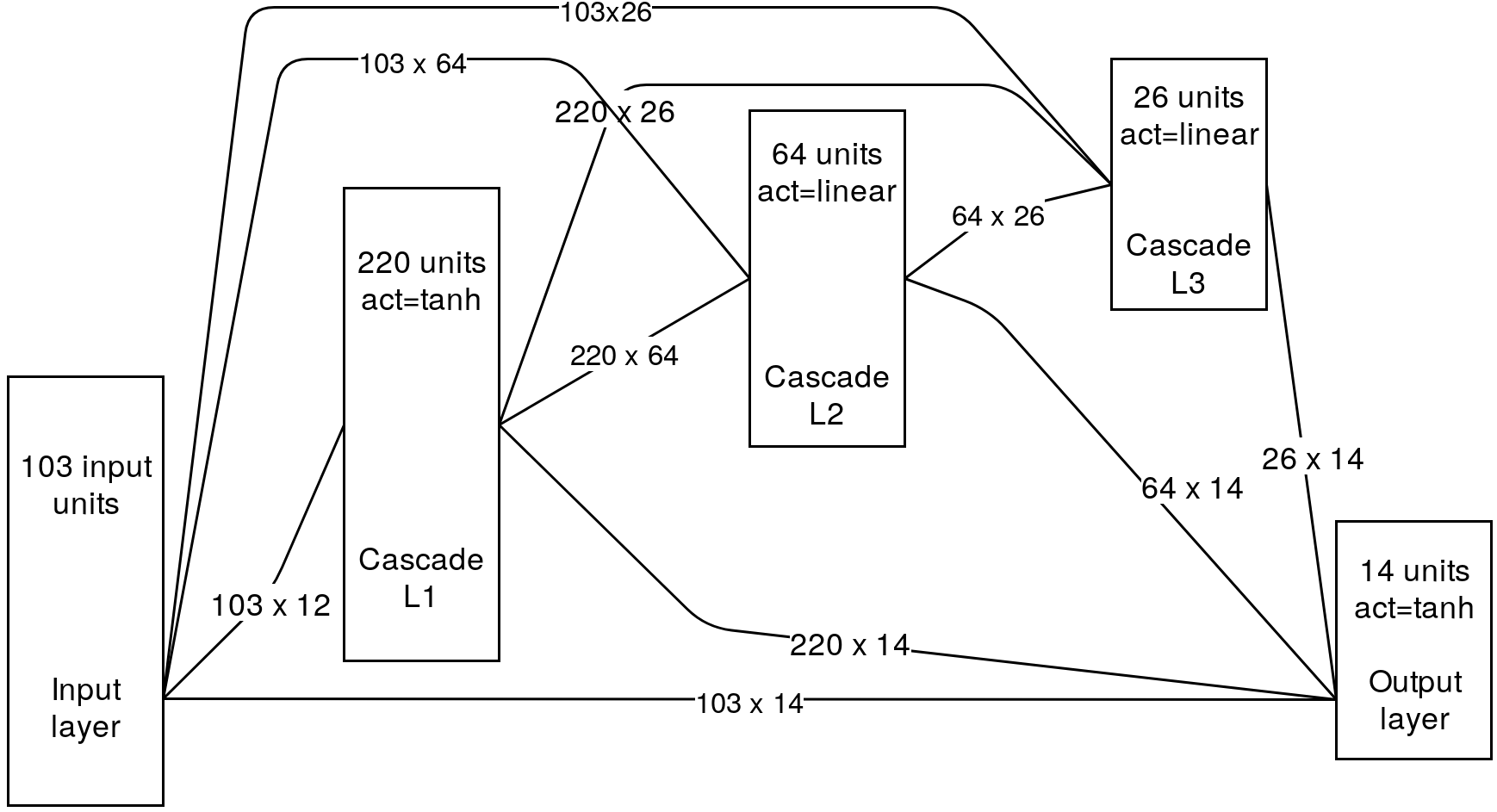}

\caption{An example network generated by a run of CascadeML on the yeast dataset. The rectangles represents layers, labelled with the number of nodes and the selected activation (act) function the layers. The lines connecting the layers indicate full connection and the text indicates the number of weights involved in the corresponding connection.\label{fig:yeast_network_example}}
\end{figure*}


A network architecture example learned by CascadeML on the yeast dataset is shown in Figure \ref{fig:yeast_network_example}. For this specific execution, three cascaded layers were selected with L1 having $220$ units and a $tanh$ activation, L2 having $64$ units and a $linear$ activation, and L3 having $26$ units and a $linear$ activation.
yeast



\section{Conclusion and Future Work\label{sec:Conclusion-and-Future}}

The work introduces a neural network algorithm, CascadeML, for multi-label
classification based on the cascade architecture, which grows the
architecture as it trains and takes label associations into account. Except setting some bounds of the hyperparameters, the
method omits the requirement of hyperparameter tuning as it automatically
determines the architecture, and uses
an adaptive first order gradient descent algorithm, iRProp-.

In an evaluation experiment CascadeML was shown to perform competitively to state-of-the-art multi-label classification algorithms, where all the other multi-label algorithms were hyperparameter tuned. CascadeML performed better on an average classifier chains, HOMER with RBF-SVM, BPMLL and MLkNN. RAkEL had the same overall averarge rank compared to CascadeML, but it did not require the extensive hyperparameter tuning.

CascadeML is the first automatic neural network algorithm with a competitive performance to hyperparameter tuned state-of-the-art multi-label classification methods, although CascadeML's performance can be improved in the cases where it does perform poorly. A limitation of the BPMLL loss function used in CascadeML is that it cannot scale with increasing number labels \cite{10.1007/978-3-662-44851-9_28}. As
the comparisons are pairwise, as the number of labels increase the
computation becomes slow like BPMLL. Therefore, it would be interesting to
investigate alternative loss functions that can still take account of label associations without the need for expensive pairwise comparisons. Also, it would be interesting to examine the patterns in which layers grow during CascadeML so as different mechanisms for adding new layers could be introduced.

\bibliographystyle{splncs04}
\bibliography{cascade_ml_draft}

\end{document}